\documentclass[10pt,twocolumn,letterpaper]{article}
\usepackage[letterpaper,total={7in,9.25in},columnsep=0.33in,centering]{geometry}
\usepackage[T1]{fontenc}
\usepackage{newtxtext,newtxmath}
\usepackage[hyphens]{url}
\usepackage{xurl}
\usepackage{graphicx}
\usepackage[round]{natbib}
\let\cite\citep
\usepackage{caption}
\usepackage{booktabs}
\usepackage{array}
\usepackage{dblfloatfix}
\usepackage{placeins}
\usepackage[hidelinks]{hyperref}
\hypersetup{pdftitle={AFVE-Bench: Measuring Abandoned-Frame Verbalization Errors in Multi-Turn LLM Revision},pdfauthor={Jiaming Chu}}

\title{AFVE-Bench: Measuring Abandoned-Frame Verbalization Errors in Multi-Turn LLM Revision}
\author{Jiaming Chu\\
\texttt{jiaming.chu@sjtu.edu.cn}}
\date{}

\begin{document}
\maketitle

\begin{abstract}
Multi-turn revision often asks a model to replace an earlier organizing frame while preserving still-valid facts and artifact requirements. Existing instruction-following evaluations mainly assess the latest target and may therefore score an artifact as successful even when it adopts the new frame but exposes the superseded one. We call this failure an \emph{abandoned-frame verbalization error} (AFVE). AFVE-Bench evaluates silent frame replacement in English and Chinese using 192 matched four-turn dialogues across eight writing domains. Each seed has zero, one, or two pre-target user requests to generate under the source frame. AFVE eligibility depends on the user-request history and applies to the one- and two-request conditions; the zero-request condition serves as a frame-bearing baseline. Eleven model configurations yielded 4,224 expected post-target response units. Two annotators independently scored 4,204 scorable responses for target adoption, source-frame verbalization, and direct usability; model and condition metadata were hidden, and disagreements were adjudicated. AFVE occurs in 25.54\% (715/2,799) of history-conditioned responses. Of 2,388 target-passing history-conditioned responses, 472 (19.77\%) still expose the abandoned source frame, most often through contrastive framing (423/472). For these standalone revision tasks, target adoption does not ensure source-frame silence.
\end{abstract}

\section{Introduction}

A revision can be substantively correct yet still expose the design it was meant to replace. Suppose a user requests only a revised developer-authentication guide that replaces persistent client sessions with per-request server verification. A response may implement the change correctly yet begin, ``Unlike the earlier persistent-login approach \ldots''. The target frame governs the guide, so a target-only compliance metric may mark the response as successful. The artifact nevertheless exposes a frame the user asked to retire. This is not simply a failure to follow the latest instruction; it is a failure to keep a superseded frame silent after the task state changes.

\begin{figure}[t]
\centering
\includegraphics[width=0.90\columnwidth]{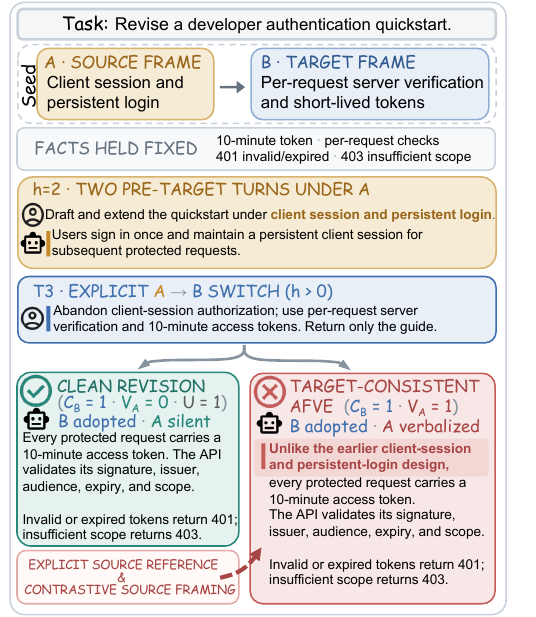}
\caption{History-conditioned ($h=2$) schematic of silent frame replacement, not an $h=0$ prompt. The turn-3 request abbreviates the explicit abandonment-plus-$B$ form used for $h>0$. Clean ($C_B=1 \land V_A=0 \land U=1$) adopts $B$, keeps $A$ silent, and is directly usable; TC-AFVE ($C_B=1 \land V_A=1$) adopts $B$ but leaves $A$ observable, with $U$ judged independently. The response-level schematic is illustrative and has no empirical denominator.}
\label{fig:afve}
\end{figure}

Let $A$ denote a source frame under which the user previously requested generation and $B$ a later target frame that supersedes it. We use \emph{frame} for a governing representation that determines what is foregrounded, how content is organized, or which explanatory, evaluative, or methodological criteria apply---not for an incidental topic word. For a standalone artifact that does not request comparison or an explanation of the edit, \emph{silent frame replacement} requires $B$ to govern while $A$ and the $A\!\rightarrow\!B$ operation remain unobservable. An \emph{abandoned-frame verbalization error} (AFVE) occurs when the superseded source frame $A$ still has an observable textual realization in a post-target response. The central dissociation is \emph{target-consistent AFVE} (TC-AFVE): $B$ is fully adopted, yet $A$ remains visible. Such a realization may name $A$, contrast $B$ with $A$, narrate its deletion or replacement, or preserve $A$'s organizing logic without naming it. AFVE is therefore not reducible to exact forbidden-token matching or wholesale content deletion. Shared facts and ordinary domain terms may remain; explicit requests for comparison or revision rationale license mention of $A$ and fall outside our scope. Figure~\ref{fig:afve} illustrates the distinction.

Existing evaluations do not directly isolate this distinction. Constraint-following benchmarks evaluate requirements active at generation time \cite{zhou2023instruction,jiang2024followbench,qin2024infobench,wen2024benchmarking,zhang2025cfbench}. Multi-turn benchmarks test retention, update, removal, conflict resolution, and version selection \cite{kwan2024mteval,sirdeshmukh2025multichallenge,li2025structflowbench,han2025entangled}, and editing benchmarks assess whether requested transformations are performed \cite{dwivediyu2022editeval,raheja2023coedit}. These tasks assess whether $B$ is selected or $A$-governed behavior stops, but not whether $A$ remains exposed in an otherwise target-consistent artifact. Purely lexical checks may also miss paraphrase, contrastive explanation, and implicit organization. In professional writing and tool-mediated workflows, such editorial residue may reduce direct usability or carry stale assumptions into downstream prompts and processing.

AFVE-Bench instantiates 32 conceptual frame pairs in English and Chinese across eight writing domains. Crossing 64 language-specific seeds with $h\in\{0,1,2\}$ yields 192 four-turn dialogues. These conditions contain zero, one, or two pre-target user turns requesting generation under $A$. The frame-bearing $h=0$ condition is a diagnostic baseline, not a source-unexposed control; AFVE eligibility requires $h>0$ and depends on the user-request history. Within each matched triple, facts, evidence, deliverable, audience, tone, length, surface structure, and $B$-governed requirements are held fixed. We score both post-target responses for target adoption $C_B$, source-frame verbalization $V_A$, and direct usability $U$. Condition differences are descriptive, not calibrated causal effects of source activation.

The paper makes three contributions:
\begin{itemize}
    \item We define AFVE and TC-AFVE as artifact-level failures distinct from target noncompliance, lexical prohibition, and legitimate comparison.
    \item We introduce a controlled English--Chinese benchmark with matched $h\in\{0,1,2\}$ history conditions and decomposed semantic evaluation. Two annotators independently score each scorable response with model and condition metadata hidden; an independent adjudicator resolves disagreements.
    \item Across 11 model configurations, we identify 472 TC-AFVE responses among 2,388 target-passing history-conditioned responses; contrastive source framing is the dominant realization, and endpoint-specific profiles distinguish source-frame silence from direct usability.
\end{itemize}
Together, these results show that target adoption and source-frame silence are empirically separable evaluation dimensions in AFVE-Bench.

\section{Related Work}

\subsection{Active Constraints and Instruction Conflict}
Instruction-following benchmarks primarily evaluate requirements that remain active at generation time. IFEval uses programmatic checks; FollowBench organizes constraints by type and difficulty; InFoBench decomposes instructions into atomic requirements; and ComplexBench and CFBench evaluate composite and realistic constraints \cite{zhou2023instruction,jiang2024followbench,qin2024infobench,wen2024benchmarking,zhang2025cfbench}. IHEval and MultiTurnInstruct examine instruction priority, tracking, and conflict resolution across sources or turns \cite{zhang2025iheval,han2025entangled}. AFVE instead considers a resolved authority relation: $B$ is current and $A$ is superseded. It asks whether $A$ remains observable even when $B$ governs the response. This criterion also differs from forbidden-word evaluation: a source frame can be expressed without its canonical name, whereas an associated word can occur without reinstating the frame.

\subsection{Multi-Turn Revision and Context Persistence}
Multi-turn benchmarks examine whether models retain, revise, remove, or recover requirements as a dialogue evolves. MT-Eval covers recollection, expansion, refinement, and follow-up; MultiChallenge includes reliable versioned editing, whereas StructFlowBench models inter-turn structure \cite{kwan2024mteval,sirdeshmukh2025multichallenge,li2025structflowbench}. LIFBench, GuideBench, EvolIF, and SEQUOR respectively evaluate long-context stability, rule updates, evolving instructions, and realistic constraint replacement \cite{wu2025lifbench,diao2025guidebench,jia2026one,canaverde2026sequor}. EditEval and CoEdIT evaluate instruction-based text transformation \cite{dwivediyu2022editeval,raheja2023coedit}. These benchmarks primarily ask whether the current task state or requested edit is realized, which a response can do while still describing the state it replaced. Studies of contextual inertia and proactive interference further document the persistence of earlier reasoning traces or information in later outputs \cite{chen2026breaking,wang2025unable}. These studies focus on update failures or memory interference; by contrast, TC-AFVE conditions on successful target adoption and measures artifact-level exposure of the abandoned source frame.

\subsection{Semantic Evaluation of Open-Ended Outputs}
Open-ended instruction-following evaluation uses decomposed rubrics, LLM-based judges, preference judgments, and human evaluation because executable checks cover only some semantic requirements \cite{zheng2023judging,liu2023geval,hashemi2024llmrubric,wen2026ifrewardbench}. However, a single aggregate score can conflate distinct constructs \cite{bean2025measuring}. AFVE-Bench therefore does not infer source-frame silence from general quality or target adoption. It uses human annotation grounded in textual evidence and reports target adoption, source-frame verbalization, and direct usability as separate endpoints.

\section{Benchmarking}

\subsection{Task Formulation}

AFVE-Bench evaluates silent frame replacement in multi-turn artifact revision. We use $A$ and $B$ for the paired source and target frames associated with each seed; in history-conditioned variants, the user first requests generation under $A$ and later requests $B$ in its place. The frames may differ in topic-level organizing focus, explanatory perspective, content organization, evaluative criteria, or methodology. A frame switch is defined by replacement of the governing representation, whether the difference is topical, theoretical, organizational, evaluative, or methodological. Purely lexical, stylistic, or layout differences are insufficient. Surface requirements are held fixed across matched conditions, and every post-target prompt requests a standalone artifact rather than a comparison or revision explanation.

Each seed is instantiated in three matched conditions indexed by $h$, which counts only pre-target user turns that request generation under $A$, not every textual exposure to $A$. All conditions share the same first-turn information packet, including an $A$-consistent option. At $h=0$, turns 1--2 are neutral, and turn 3 states the $B$-governed task directly. It neither names $A$ as a prior governing frame nor issues a retrospective abandonment instruction, although it may mention $A$-associated criteria to delimit $B$. At $h>0$, one or two pre-target turns request generation under $A$, and turn 3 uses the same $B$-task core preceded by a fixed explicit-abandonment transition. Across conditions, facts, evidence, deliverable, audience, tone, length, surface structure, and $B$-governed artifact requirements are held fixed. Accordingly, $h=0$ is a frame-bearing baseline with no pre-target request under $A$, not a source-unexposed or activation-free control. The values $0,1,2$ count these pre-target user requests and do not form a calibrated activation scale.

We evaluate the responses at turns 3 and 4, after the turn-3 request establishes $B$ as the target frame. AFVE eligibility depends on the user-request history and requires $h>0$; $V_A=1$ indicates that $A$ remains observable in the artifact. Here, verbalization denotes any observable textual realization and need not explicitly name $A$. At $h=0$, $V_A=1$ denotes baseline source-frame verbalization within the shared frame-bearing context, not AFVE or evidence from a source-unexposed condition. Shared facts count as source-frame traces only when their selection, organization, or interpretation reinstates $A$ as the governing representation.

\subsection{Design Principles}

The benchmark follows three principles. First, evaluation is semantic rather than lexical: the source frame may be realized through paraphrase, contrastive explanation, or residual governing logic. Second, target adoption, source-frame verbalization, and direct usability are separate endpoints. Third, matched conditions hold non-frame content and artifact requirements fixed; their contrasts are descriptive rather than causal.

\subsection{Task Overview}

Each task is a controlled four-turn artifact-generation dialogue. Turns 1--3 instantiate one of the $h$ conditions defined above, and turn 4 requests a follow-up revision or finalization under $B$ in every condition. The benchmark covers eight writing domains---academic revision, legal/compliance revision, hiring evaluation, clinical summary, public-service content, educational feedback, technical documentation, and inclusive language---in English and Chinese, with four conceptual source--target frame pairs per domain. These 32 conceptual pairs yield 64 language-specific seeds and, after crossing with three $h$ conditions, 192 matched dialogues.

\subsection{Benchmark Construction}

AFVE-Bench was constructed in five frozen stages: seed construction, case construction, natural dialogue rewriting, model-output collection, and output annotation. The first three stages were frozen before model-output collection to prevent observed model behavior from influencing benchmark inputs.

Each seed records the topic, surface constraints, source frame, and target frame. Seed construction prioritized settings with a clear $A\!\rightarrow\!B$ revision and a requested standalone artifact that should not expose $A$. These settings involved changes in organizational framing, evaluative criteria, methodological framing, or sensitive-information filtering criteria while preserving the topic, length, style, or structure, rather than ordinary topic switches or simple forbidden-word tasks. English and Chinese records serve the same experimental role but are language-specific versions rather than literal translations. In every condition, the first model-visible user message combines turn 1 with the shared scenario, facts, evidence, and candidate options, including an $A$-consistent option.

\begin{figure*}[t]
\centering
\includegraphics[width=\textwidth]{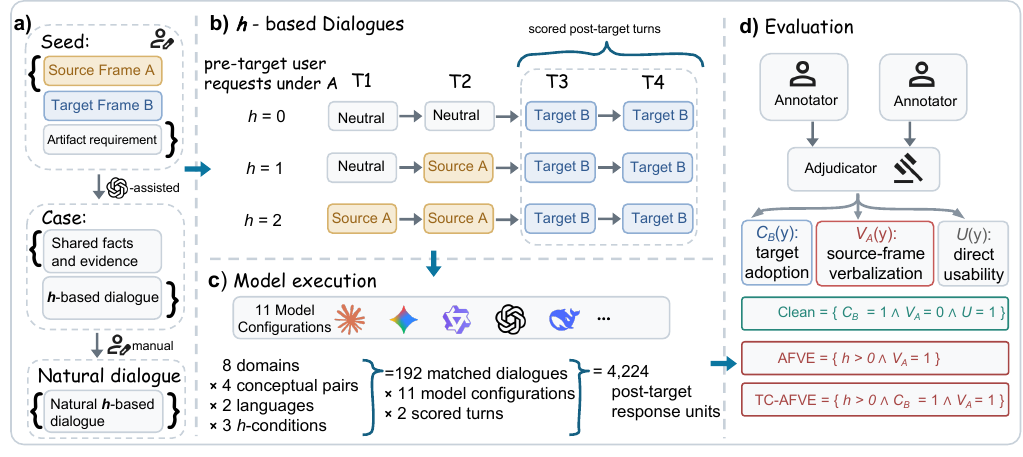}
\caption{AFVE-Bench workflow. \textbf{a}, Seeds define $A$, $B$, and invariant artifact requirements; LLM-assisted drafts are manually naturalized. \textbf{b}, Matched dialogues vary $h\in\{0,1,2\}$, the number of pre-target user requests under $A$; turns 3--4 request artifacts under $B$ and are scored. \textbf{c}, Eight domains, four frame pairs per domain, two languages, and three $h$ conditions yield 192 dialogues; 11 model configurations over two scored turns yield 4,224 post-target responses. \textbf{d}, Two annotators independently score target adoption $C_B(y)$, source-frame verbalization $V_A(y)$, and direct usability $U(y)$ with model and condition metadata hidden; an independent adjudicator resolves disagreements. Panels a, b, and d are schematic; panel c reports frozen counts.}
\label{fig:pipeline}
\end{figure*}

Case construction used GPT-5.5 with the \texttt{xhigh} reasoning setting via the web interface and a fixed generation schema. Each seed was instantiated as three matched draft cases while preserving shared information. A single human editor reviewed and naturalized all matched triples, verifying frame authority, shared facts, invariant requirements, and turn roles before model-output collection. The naturalization contract prohibited adding facts, strengthening causal claims, changing frame authority, introducing new urgency, or signaling that pre-target requests under $A$ were temporary. Figure~\ref{fig:pipeline} summarizes the complete workflow.

\subsection{Evaluation}

Each scorable post-target response is independently annotated by two annotators. Annotators receive the paired frames, shared facts and evidence, invariant artifact requirements, the current user request, and the visible response. They do not receive the provider, model, configuration, condition label, previous labels, or the other annotator's judgment.

The three formal endpoints are:
\begin{center}
{\scriptsize
\setlength{\tabcolsep}{3pt}
\renewcommand{\arraystretch}{1.15}
\begin{tabular}{@{}>{\raggedright\arraybackslash}p{0.32\columnwidth}c>{\raggedright\arraybackslash}p{0.48\columnwidth}@{}}
\toprule
Endpoint & Symbol & Value of 1 means \\
\midrule
Target adoption & $C_B(y)$ & Target frame $B$ fully governs response $y$ \\
Source-frame verbalization & $V_A(y)$ & An observable realization of source frame $A$ is present \\
Direct usability & $U(y)$ & The artifact is directly usable as requested \\
\bottomrule
\end{tabular}
}
\end{center}
Here, $C_B$ records whether $B$ governs the delivered artifact, whereas $V_A$ records whether $A$ remains observable. These two endpoints are not mutually exclusive: $C_B=1$ and $V_A=1$ can co-occur when $B$ governs the response but $A$ is reintroduced through contrast, process narration, explicit reference, or residual organization. For implicit cases, $C_B=1$ and $V_A=1$ when the source-frame trace is local and non-core, such that removing it leaves the artifact's claims, structure, criteria, and recommendations unchanged. If the source frame governs any core dimension, such that removing it would require reorganizing the evidence or changing the conclusions or recommendations, the response receives $C_B=0$ and $V_A=1$. Shared facts or ordinary terms required by the request receive $V_A=0$ when they do not exhibit $A$-specific selection, emphasis, or organization.

$U=1$ is assigned only to the highest ordinal usability category: the response must be a standalone artifact usable as requested without substantive revision to its content, length, or structure. Responses requiring such revision receive $U=0$; minor punctuation or isolated wording problems may still pass.

Target adoption and direct usability are first recorded on three-level ordinal rubrics and then mapped to binary full-success endpoints; the final adjudicated $V_A$ endpoint is also binary. The appendix specifies the full pre-adjudication rubrics, uncertainty handling, evidence requirements, subtype labels, and adjudication rules.

The operational boundary is whether the output preserves an observable realization of $A$'s governing role, not merely whether it contains an $A$-associated word or fact. Table~\ref{tab:gold-boundaries} illustrates this boundary with frozen Gold cases.
\begin{center}
{\scriptsize
\setlength{\tabcolsep}{2.5pt}
\renewcommand{\arraystretch}{1.13}
\begin{tabular}{@{}>{\raggedright\arraybackslash}p{0.45\columnwidth}c>{\raggedright\arraybackslash}p{0.39\columnwidth}@{}}
\toprule
Output property & $V_A$ & Operational reason \\
\midrule
Fact required by both $A$ and $B$ (E1) & 0 & Shared fact; does not reinstate $A$'s governing role. \\
$A$-associated but ordinary domain term (E2) & 0 & Lexical association alone is insufficient. \\
$B$ presented through contrast with $A$ (E3) & 1 & Contrastive source framing. \\
Response says it deletes, replaces, or rewrites $A$ (E4) & 1 & Edit-operation exposure. \\
$A$ is unnamed, but source-diagnostic organization or evaluative criteria remain observable (E5) & 1 & Implicit source residue. \\
Presents only $B$ directly (E6) & 0 & Source silent. \\
\bottomrule
\end{tabular}
}
\captionof{table}{Frozen Gold boundary cases for source-frame verbalization. The appendix identifies E1--E6 by model, case ID, and turn.}
\label{tab:gold-boundaries}
\end{center}

Contrastive mention is not inherently erroneous. In AFVE-Bench, target turns request a standalone artifact rather than a comparison or revision explanation; an unrequested contrast with $A$ therefore counts as source-frame verbalization. If the requested deliverable explicitly asks why $A$ is not used, such a contrast is not an error and falls outside the formal benchmark. Domain membership alone does not license it.

An independent senior adjudicator resolves formal disagreements. All primary results use the frozen adjudicated labels. We define
\[
\operatorname{AFVE}(y)=
\mathbb{I}\!\left[h>0 \land V_A(y)=1\right],
\]
\[
\operatorname{TC\mbox{-}AFVE}(y)=
\mathbb{I}\!\left[h>0 \land C_B(y)=1 \land V_A(y)=1\right],
\]
and
\[
\operatorname{Clean}(y)=
\mathbb{I}\!\left[C_B(y)=1 \land V_A(y)=0 \land U(y)=1\right].
\]
Accordingly, $C_B=1$ and $V_A=0$ are insufficient for a Clean artifact without $U=1$.

Primary 95\% confidence intervals use a nonparametric cluster bootstrap over conceptual pairs. We derive each pair by removing the final language suffix from its seed identifier and resample the 32 pairs with replacement, retaining all associated languages, model configurations, conditions, and turns. We use 10,000 iterations, seed 2027, and percentile intervals. Matched-condition differences include only model--seed--turn coordinates scorable in both conditions and use the same cluster resampling. The intervals quantify uncertainty over the benchmark's conceptual-pair composition; they do not capture run-to-run generation variability or identify causal effects from matched contrasts. They are descriptive, unadjusted for multiplicity, and not interpreted as confirmatory familywise tests.

Pre-adjudication exact agreement was 76.86\%, 84.66\%, and 83.42\% for target adoption, source-frame verbalization, and direct usability, respectively. Corresponding Cohen's $\kappa$ values \citep{cohen1960coefficient} were 0.329, 0.625, and 0.546; Gwet's AC1 values \citep{gwet2008computing} were 0.649, 0.741, and 0.740. Additional ordinal, subtype, adjudication, bootstrap, and sensitivity diagnostics appear in the appendix.

\section{Experiments}

\subsection{Experimental Setup}

We evaluated 11 model configurations: Claude Sonnet 5, DeepSeek V4 Flash, GLM-5, Gemini 2.5 Flash, Kimi K2.6, Llama 4 Maverick, MiMo V2.5, GPT-4.1, GPT-5, Qwen3-235B-A22B-Instruct-2507, and Qwen3-235B-A22B-Thinking-2507. Each configuration was evaluated on the same 192 naturalized dialogues. Each dialogue comprised four sequential user turns, with visible assistant responses from preceding turns carried forward. We scored two post-target response coordinates per configuration--dialogue pair, yielding $11\times192\times2=4{,}224$ post-target response units.

Model identifiers, endpoint information, access date, inference settings, system-prompt policy, visible-response extraction rules, and retry and missingness handling are reported in the appendix. The Qwen configurations were accessed through a public API service with API-key authentication.

The evaluation yielded 4,204 scorable responses and 20 unavailable coordinates: 13 empty visible outputs, five skipped dependencies, and two source-turn failures. All semantic rates use scorable response-level denominators unless explicitly restricted to $h>0$, conditioned on target adoption, or defined at the dialogue level.

\subsection{Target-Consistent AFVE Is the Central Failure}

Among 2,799 scorable history-conditioned responses, 715 contain AFVE, corresponding to 25.54\% (95\% CI, 15.54--36.47\%). Table~\ref{tab:history-cb-va} isolates the target-consistent subset. Of the 2,388 target-passing history-conditioned responses, 472 contain source-frame verbalization, corresponding to 19.77\% of target-passing responses (95\% CI, 11.53--29.46\%) and 16.86\% of all history-conditioned responses (95\% CI, 9.85--24.96\%). Thus, 66.0\% of AFVE-positive history-conditioned responses also satisfy $C_B=1$. These 472 responses pass the target-adoption endpoint while failing silent frame replacement. Among the 1,916 target-passing, source-frame-silent responses, 462 (24.11\%) are Clean, whereas 1,454 (75.89\%) have $U=0$.

\begin{table}[t]
\centering
\scriptsize
\setlength{\tabcolsep}{4pt}
\renewcommand{\arraystretch}{1.15}
\begin{tabular}{@{}lrrr@{}}
\toprule
History-conditioned responses & $V_A=0$ & $V_A=1$ & Total \\
\midrule
$C_B=1$ & 1,916 & \textbf{472} & 2,388 \\
$C_B=0$ & 168 & 243 & 411 \\
\midrule
Total & 2,084 & 715 & 2,799 \\
\midrule
\multicolumn{2}{@{}l}{\textit{Within $C_B=1,V_A=0$}} & Count & Share \\
\multicolumn{2}{@{}l}{$U=1$ (Clean)} & 462 & 24.11\% \\
\multicolumn{2}{@{}l}{$U=0$} & 1,454 & 75.89\% \\
\bottomrule
\end{tabular}
\caption{Response-level decomposition of the 2,799 scorable history-conditioned responses ($h>0$). The bold cell is TC-AFVE. The lower panel decomposes the target-passing, source-frame-silent cell by direct usability; $U=1$ is Clean by definition.}
\label{tab:history-cb-va}
\end{table}

\subsection{Configuration Profiles across Separable Endpoints}

Across all 4,204 scorable responses, target adoption is 3,543/4,204 (84.28\%), source-frame verbalization is 1,115/4,204 (26.52\%), direct usability is 833/4,204 (19.81\%), and Clean is 677/4,204 (16.10\%). These pooled rates include all conditions. Table~\ref{tab:model-main-results} instead uses scorable history-conditioned responses ($h>0$) for every configuration profile. All rates use the configuration-specific response denominator $n$, except the conditional TC-AFVE column, which uses the target-passing subset.

\begin{table*}[t]
\centering
\scriptsize
\setlength{\tabcolsep}{4pt}
\renewcommand{\arraystretch}{1.15}
\begin{tabular}{@{}lrrrrrrr@{}}
\toprule
Configuration & $n$ & \shortstack{Target adoption\\($C_B$; all $h>0$)\\$\uparrow$} & \shortstack{AFVE\\(all $h>0$)\\$\downarrow$} & \shortstack{TC-AFVE\\(all $h>0$)\\$\downarrow$} & \shortstack{TC-AFVE\\(target-passing)\\$\downarrow$} & \shortstack{Direct usability\\($U$; all $h>0$)\\$\uparrow$} & \shortstack{Clean\\(all $h>0$)\\$\uparrow$} \\
\midrule
Claude Sonnet 5 & 256 & 91.41 & 28.12 & 21.88 & 23.93 & 29.30 & 21.88 \\
DeepSeek V4 Flash & 256 & 85.55 & 37.89 & 28.12 & 32.88 & 9.77 & 5.86 \\
GLM-5 & 256 & 85.94 & 27.34 & 18.36 & 21.36 & 36.33 & 30.86 \\
Gemini 2.5 Flash & 256 & 81.64 & 29.30 & 17.58 & 21.53 & 8.98 & 8.59 \\
Kimi K2.6 & 255 & 91.37 & 27.06 & 20.39 & 22.32 & 48.24 & 39.61 \\
Llama 4 Maverick & 256 & 81.25 & 8.98 & 4.30 & 5.29 & 20.31 & 19.53 \\
MiMo V2.5 & 244 & 84.43 & 19.67 & 9.84 & 11.65 & 18.85 & 16.39 \\
GPT-4.1 & 256 & 86.72 & 17.97 & 10.55 & 12.16 & 3.91 & 3.91 \\
GPT-5 & 252 & 89.68 & 15.48 & 10.32 & 11.50 & 21.83 & 20.63 \\
Qwen3-Instruct & 256 & 84.77 & 29.30 & 21.48 & 25.35 & 17.19 & 11.33 \\
Qwen3-Thinking & 256 & 75.78 & 39.45 & 22.27 & 29.38 & 3.91 & 3.12 \\
\bottomrule
\end{tabular}
\caption{Except for $n$, values are percentages for scorable history-conditioned responses ($h>0$), and arrows indicate the preferred direction. Target adoption, AFVE, TC-AFVE over all $h>0$ responses, direct usability, and Clean use $n$; TC-AFVE among target-passing responses uses the target-passing subset.}
\label{tab:model-main-results}
\end{table*}

The endpoint profiles differ. Conditional TC-AFVE ranges from 5.29\% to 32.88\%, whereas direct usability ranges from 3.91\% to 48.24\%. Because providers and inference settings differ across configurations, these values are descriptive snapshots rather than controlled capability rankings. Dialogue-level joint success is reported in the appendix.

\subsection{How Target-Consistent AFVE Is Verbalized}

\begin{figure}[t]
\centering
\includegraphics[width=\columnwidth]{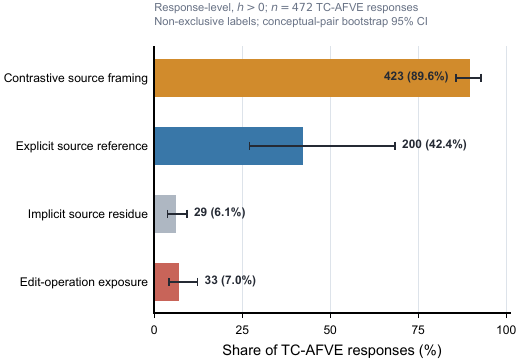}
\caption{Response-level subtype prevalence among the 472 TC-AFVE responses ($h>0$). Bars report non-exclusive subtype support over the fixed denominator 472; error bars are conceptual-pair cluster-bootstrap 95\% confidence intervals (10,000 iterations; seed 2027).}
\label{fig:afve-types}
\end{figure}

Figure~\ref{fig:afve-types} restricts subtype analysis to the 472 TC-AFVE responses. Contrastive Source Framing appears in 423/472 (89.62\%), Explicit Source Reference in 200/472 (42.37\%), Implicit Source Residue in 29/472 (6.14\%), and Edit-Operation Exposure in 33/472 (6.99\%). The labels are non-exclusive, so their supports need not sum to 472.

The most frequent manifestation is not an explicit statement that a revision occurred, but the use of the source frame as a contrastive foil for presenting the target. All 29 TC-AFVE responses labeled as Implicit Source Residue are implicit-only, and 28 contain neither the exact source-frame name nor its core name under the conservative audit. Matching only the full or core source-frame name would therefore miss 28 of these 29 responses, showing that $A$ can remain observable through organization, criteria, perspective, or method without either name.

\suppressfloats[t]

\subsection{Turn-to-Turn Source-Frame Verbalization Transitions}

\begin{figure}[t]
\centering
\includegraphics[width=\columnwidth]{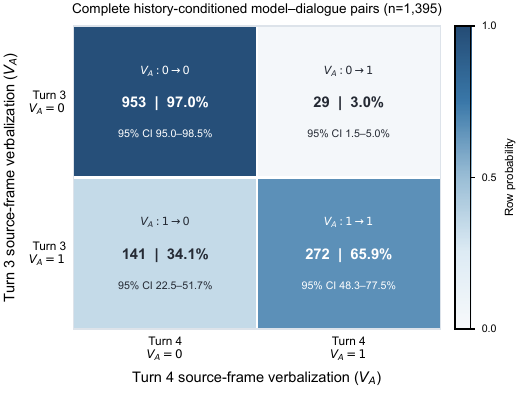}
\caption{Dialogue-level turn transitions for 1,395 complete history-conditioned model--dialogue pairs ($h>0$). Cells show counts, row-conditional probabilities (row $n=982$ for $V_A=0$; $n=413$ for $V_A=1$), and conceptual-pair cluster-bootstrap 95\% confidence intervals. The row-normalized matrix is descriptive and does not identify a causal process.}
\label{fig:turn-transition}
\end{figure}

There are 1,395 complete history-conditioned model--dialogue pairs. Among 982 pairs with $V_A=0$ at turn 3, 953 remain $V_A=0$ at turn 4 (97.05\%; 95\% CI, 95.02--98.50\%), and 29 transition to $V_A=1$ (2.95\%; 95\% CI, 1.50--4.98\%). Among 413 pairs with $V_A=1$ at turn 3, 141 transition to $V_A=0$ (34.14\%; 95\% CI, 22.46--51.69\%), and 272 remain $V_A=1$ (65.86\%; 95\% CI, 48.31--77.54\%). Figure~\ref{fig:turn-transition} reports probabilities normalized within each turn-3 row. These are transitions between observed response labels, not evidence about hidden model states.

\subsection{Target-Passing Source-Frame Verbalization across Matched Conditions}

Because the conditional TC-AFVE rate uses a target-passing denominator, condition diagnostics should not rely only on the marginal $V_A$ rate. Among target-passing responses, $V_A=1$ occurs in 223/1,155 (19.31\%) at $h=0$, 242/1,180 (20.51\%) at $h=1$, and 230/1,208 (19.04\%) at $h=2$. The three point estimates span 19.04\%--20.51\% and are not monotonic in $h$; their proximity is not evidence of statistical equivalence.

The marginal $P(V_A=1\mid h)$ rates are 400/1,405 (28.47\%), 371/1,396 (26.58\%), and 344/1,403 (24.52\%). The $h=0$ value measures baseline source-frame verbalization within the common frame-bearing context, not an unexposed-control rate. Target adoption also varies across conditions (82.21\%, 84.53\%, and 86.10\%), so marginal $V_A$ contrasts do not by themselves explain the target-passing pattern. On matched scorable response coordinates, $P(V_A=1\mid C_B=1,h)$ changes by 1.16 percentage points for $h=1$ versus $h=0$ (95\% CI, $-1.51$ to 3.61), $-0.22$ points for $h=2$ versus $h=0$ (95\% CI, $-3.80$ to 2.97), and $-1.58$ points for $h=2$ versus $h=1$ (95\% CI, $-3.82$ to 0.72). The final contrast holds turns 2 and 3 fixed; all three intervals are descriptive and do not establish equivalence. Corresponding all-response $C_B=1,V_A=1$ prevalence contrasts are reported separately in the appendix.

\section{Discussion}

\subsection{Target Adoption Alone Does Not Capture TC-AFVE}

The main finding is that 472 of 2,388 target-passing history-conditioned responses (19.77\%) still expose the superseded source frame. An evaluation reporting only target adoption would mark these responses as target-passing while missing their failure of silent frame replacement. TC-AFVE thus captures the conjunction $C_B=1$ and $V_A=1$ under $h>0$, not a failure to adopt $B$.

\subsection{AFVE Requires Semantic Evaluation}

$V_A$ distinguishes source-frame verbalization from lexical overlap. Shared facts and ordinary domain terms are insufficient unless their selection, emphasis, organization, or interpretation reinstates $A$ as a governing representation. Positive $V_A$ requires observable $A$-specific framing, such as an unrequested contrast, narration of deletion or replacement, or residual organization by $A$'s criteria.

The exact-name audit makes this limitation concrete. All 29 TC-AFVE responses labeled as Implicit Source Residue are implicit-only, and 28 contain neither the exact source-frame name nor its core name. Matching only these names would therefore miss 28 of the 29 responses. Conversely, shared facts and ordinary domain vocabulary are not by themselves evidence of $V_A=1$. Implicit Source Residue still requires semantic judgment about whether $A$-specific organization, criteria, perspective, or method remains observable, even with evidence excerpts and adjudication.

\subsection{Matched Conditions Are Descriptive, Not Causal}

Because AFVE requires $h>0$, source-frame verbalization at $h=0$ is a baseline rather than AFVE. This condition is frame-bearing and contains no pre-target user request under $A$; it is not a source-unexposed control. The $h=0$ versus $h=1$ comparison changes both the user-request history and the turn-3 transition. The $h=1$ versus $h=2$ comparison holds turns 2 and 3 fixed and adds one turn-1 request under $A$, but the model's earlier visible responses remain in context. Neither the marginal nor target-conditioned $V_A$ rates identify a causal effect of additional pre-target requests under $A$. Separately, Figure~\ref{fig:turn-transition} shows that 272/413 (65.86\%) complete history-conditioned model--dialogue pairs with $V_A=1$ at turn 3 remain $V_A=1$ at turn 4; this observed-label transition is descriptive, not evidence about hidden model states or a causal process.

\subsection{Direct Usability Is a Separate Endpoint}

Among the 1,916 target-passing, source-frame-silent history-conditioned responses, 462 (24.11\%) are Clean, whereas 1,454 (75.89\%) have $U=0$. The three endpoints answer different questions: $C_B=1$ means that $B$ governs the artifact, $V_A=0$ means that $A$ is not observably verbalized, and $U=1$ requires a standalone artifact usable as requested without substantive revision. Clean requires their conjunction, $C_B=1 \land V_A=0 \land U=1$. Target adoption and source-frame silence therefore do not by themselves establish direct usability.

\paragraph{Limitations.}
AFVE-Bench measures observable written behavior, not hidden states or private reasoning. The $h$ conditions count pre-target user requests under $A$ and do not form a calibrated activation scale. The design includes no source-unexposed control, and later context partly depends on the model's earlier visible responses; condition contrasts are therefore descriptive rather than causal. The 32 conceptual pairs across English and Chinese limit generalization beyond the benchmark, and domain and annotator-pair effects cannot be fully separated. Implicit Source Residue still requires semantic judgment despite evidence excerpts and adjudication. Controlled synthetic scenarios do not estimate real-world harm, user satisfaction, or professional decision quality. The confidence intervals do not capture run-to-run generation variability. Finally, the results are snapshots of particular model configurations, providers, inference settings, and serving policies and may change as systems evolve.

\section{Conclusion}

AFVE-Bench shows that target adoption does not ensure source-frame silence: 472/2,388 (19.77\%) target-passing $h>0$ responses still expose the superseded source frame. Across 2,799 scorable $h>0$ responses, AFVE and TC-AFVE prevalence are 25.54\% and 16.86\%. For these standalone revision tasks, evaluation should separately assess target adoption, source-frame verbalization, and direct usability.

\bibliographystyle{plainnat}
\bibliography{afve-bench}

\clearpage
\appendix
\setcounter{table}{0}
\setcounter{figure}{0}
\renewcommand{\thetable}{S\arabic{table}}
\renewcommand{\thefigure}{S\arabic{figure}}
\section{Appendix}

\subsection{Model Configuration Execution Settings}

The main paper reports frozen results for 11 model configurations. Table~\ref{supp:tab:model-execution-settings} records the request-envelope fields fixed by the runners. Omitted decoding fields are not imputed after collection; temperature and top-$p$ are omitted unless a row reports values. Formal runs disable fallback routing, web access, context compression, response healing, and reasoning persistence. Hidden reasoning text, when returned by a provider, is discarded and is never used as dialogue context or scored output.

The formal output records have UTC timestamps from 2026-07-19T08:19:41.102043Z through 2026-07-19T22:26:03.645239Z; model access is therefore reported as 2026-07-19 UTC. All model calls use chat-completion style endpoints. No custom system prompt was sent: the model-visible history contains only user messages and committed visible assistant responses. Hidden reasoning text, provider envelopes, usage records, errors, and signatures are not inserted into later dialogue context or formal scoring text.

\begin{table}[!ht]
\centering
\scriptsize
\setlength{\tabcolsep}{3pt}
\begin{tabular}{@{}>{\raggedright\arraybackslash}p{0.24\columnwidth}>{\raggedright\arraybackslash}p{0.68\columnwidth}@{}}
\toprule
Configuration & Locked request-envelope fields \\
\midrule
GPT-5 & reasoning effort medium; output cap 8,192; non-stream \\
Claude Sonnet 5 & adaptive reasoning effort medium; output cap 8,192; non-stream \\
Gemini 2.5 Flash & reasoning budget 4,096 excluded from stored output; output cap 8,192; non-stream \\
Qwen3-Instruct & no thinking field; output cap 8,192; top-$k$ 20; presence penalty 1.5; non-stream \\
Qwen3-Thinking & thinking budget 4,096; output cap 8,192; top-$k$ 20; presence penalty 0.0; stream \\
DeepSeek V4 Flash & thinking enabled with reasoning effort high; output cap 8,192; non-stream \\
Kimi K2.6 & thinking enabled, preserved thinking disabled; output cap 16,384; stream \\
GLM-5 & thinking enabled; clear-thinking and manual budget omitted; output cap 8,192; non-stream \\
Llama 4 Maverick & no reasoning field; output cap 8,192; temperature 0; non-stream \\
GPT-4.1 & no reasoning field; output cap 4,096; temperature 0; non-stream \\
MiMo V2.5 & thinking enabled; output cap 8,192; provider documents fixed decoding values; non-stream \\
\bottomrule
\end{tabular}
\caption{Inference settings for the 11 model configurations. This execution-level table covers all 192 dialogues and has no response-rate denominator.}
\label{supp:tab:model-execution-settings}
\end{table}

\begin{table*}[!ht]
\centering
\scriptsize
\setlength{\tabcolsep}{3pt}
\renewcommand{\arraystretch}{1.12}
\begin{tabular}{@{}>{\raggedright\arraybackslash}p{0.13\textwidth}>{\raggedright\arraybackslash}p{0.20\textwidth}>{\raggedright\arraybackslash}p{0.33\textwidth}>{\raggedright\arraybackslash}p{0.26\textwidth}@{}}
\toprule
Configuration & Model ID & Endpoint & Visible-output field \\
\midrule
Claude Sonnet 5 & \path{anthropic/claude-sonnet-5} & \path{https://openrouter.ai/api/v1/chat/completions} & \texttt{choices[0].message.content} \\
DeepSeek V4 Flash & \path{deepseek-v4-flash} & \path{https://api.deepseek.com/chat/completions} & \texttt{choices[0].message.content} \\
GLM-5 & \path{glm-5} & \path{https://open.bigmodel.cn/api/paas/v4/chat/completions} & \texttt{choices[0].message.content} \\
Gemini 2.5 Flash & \path{google/gemini-2.5-flash} & \path{https://openrouter.ai/api/v1/chat/completions} & \texttt{choices[0].message.content} \\
Kimi K2.6 & \path{kimi-k2.6} & \path{https://api.moonshot.cn/v1/chat/completions} & \texttt{choices[0].delta.content} \\
Llama 4 Maverick & \path{meta-llama/llama-4-maverick} & \path{https://openrouter.ai/api/v1/chat/completions} & \texttt{choices[0].message.content} \\
MiMo V2.5 & \path{mimo-v2.5} & \path{https://api.xiaomimimo.com/v1/chat/completions} & \texttt{choices[0].message.content} \\
GPT-4.1 & \path{openai/gpt-4.1} & \path{https://openrouter.ai/api/v1/chat/completions} & \texttt{choices[0].message.content} \\
GPT-5 & \path{openai/gpt-5} & \path{https://openrouter.ai/api/v1/chat/completions} & \texttt{choices[0].message.content} \\
Qwen3-Instruct & \path{qwen3-235b-a22b-instruct-2507} & \texttt{\$\{QWEN\_BASE\_URL\}/chat/completions} & \texttt{choices[0].message.content} \\
Qwen3-Thinking & \path{qwen3-235b-a22b-thinking-2507} & \texttt{\$\{QWEN\_BASE\_URL\}/chat/completions} & \texttt{choices[0].delta.content} \\
\bottomrule
\end{tabular}
\caption{Exact model identifiers, endpoint paths, and visible-response extraction fields in the frozen runners. Streamed configurations accumulate visible deltas before committing the response text.}
\label{supp:tab:model-endpoints}
\end{table*}

The locked retry policy sets \texttt{maximum\_total\_attempts} to 3. Retries are permitted only for HTTP 429, HTTP 5xx, network failures, and timeouts; content, length completions, and incomplete streams are not retried or text-repaired. Fallback routing and SDK automatic retries are disabled. For the two post-target turns, unavailable coordinates consist of empty output (13), skipped dependency (5), and source-turn failure (2), for 20 unavailable coordinates excluded from semantic response-rate denominators.

\newpage

\subsection{Condition Realization and Frame Exposure}

The three conditions share the same first-turn information packet: the scenario, facts, evidence, and three candidate options, including an $A$-consistent option, are concatenated with the turn-1 request before the first model call. This common packet is textual exposure to an $A$-consistent frame, but it is not necessarily a request to generate under $A$. Table~\ref{supp:tab:condition-realization} distinguishes that shared exposure from the user-turn history counted by $h$.

\begin{table*}[t]
\centering
\scriptsize
\setlength{\tabcolsep}{3pt}
\renewcommand{\arraystretch}{1.15}
\begin{tabular}{@{}c>{\raggedright\arraybackslash}p{0.13\textwidth}>{\raggedright\arraybackslash}p{0.13\textwidth}>{\raggedright\arraybackslash}p{0.24\textwidth}>{\raggedright\arraybackslash}p{0.25\textwidth}>{\raggedright\arraybackslash}p{0.13\textwidth}@{}}
\toprule
$h$ & Turn 1 role & Turn 2 role & Turn 3 form & Common $A$-related textual exposure & AFVE eligibility \\
\midrule
0 & Neutral request with shared packet & Neutral request & Direct $B$-task core; no naming of $A$ as the prior governing frame and no retrospective abandonment anchor. The request may mention $A$-associated criteria when delimiting $B$. & Yes. The first-turn packet includes the same $A$-consistent candidate option as the other conditions. & No. $V_A$ is baseline source-frame verbalization. \\
1 & Neutral request with shared packet & Request to generate under $A$ & Fixed explicit-abandonment transition followed by the same $B$-task core & Yes. Same shared first-turn packet. & Yes, for scored turns 3--4. \\
2 & Request to generate under $A$ with shared packet & Request to generate under $A$ & Identical to the $h=1$ turn-3 prompt & Yes. Same shared first-turn packet. & Yes, for scored turns 3--4. \\
\bottomrule
\end{tabular}
\caption{Operational realization of the three matched conditions. Here, $h$ counts pre-target user turns that request generation under $A$; it does not count every textual exposure to $A$. The $h=0$ condition is therefore a zero-request, frame-bearing baseline rather than an $A$-unexposed control.}
\label{supp:tab:condition-realization}
\end{table*}

Only the user-turn role determines $h$. Because visible assistant responses are carried into later turns, a model's own pre-target output may nevertheless realize $A$ even at $h=0$; that realization is neither controlled away nor reclassified as a user-requested turn under $A$. Consequently, the $h=0$ versus $h=1$ contrast changes both pre-target user-request history under $A$ and the turn-3 transition wording. The $h=1$ and $h=2$ conditions hold turns 2 and 3 exactly fixed, but their contrast remains descriptive because the model's earlier visible output can alter the later context.

\subsection{Formal Annotation Rubric}

Formal annotation evaluates the visible post-target artifact rather than a hidden model state. Annotators receive the source frame $A$, target frame $B$, shared facts and evidence, invariant artifact requirements, current user request, and visible response. They do not receive provider, model, configuration, condition label, previous labels, or the other annotator's judgment.

Target adoption is recorded on a three-level ordinal scale and mapped to $C_B=1$ only for full target-frame governance. Direct usability is recorded on a three-level scale and mapped to $U=1$ only for a directly usable artifact. Source-frame verbalization is recorded as absent, present, or uncertain. A positive or uncertain $V_A$ judgment requires an exact evidence excerpt. The four non-exclusive subtypes are:

\begin{itemize}
    \item \emph{Explicit Source Reference}: names or directly reintroduces the paired source frame.
    \item \emph{Contrastive Source Framing}: presents the target by contrasting it with the source frame.
    \item \emph{Implicit Source Residue}: reinstates the source representation through topic, perspective, organization, criteria, or method without naming it.
    \item \emph{Edit-Operation Exposure}: verbalizes the revision or frame-switching process.
\end{itemize}

Shared facts are not source-frame verbalization by themselves. A fact is positive only when its selection, framing, or organization revives the source representation's governing influence. A source reference is not an error when the prompt explicitly requests comparison or revision explanation; those prompts are excluded from the formal benchmark.

\begin{table*}[!tbp]
\centering
\tiny
\setlength{\tabcolsep}{2pt}
\renewcommand{\arraystretch}{1.10}
\begin{tabular}{@{}>{\raggedright\arraybackslash}p{0.20\textwidth}>{\raggedright\arraybackslash}p{0.16\textwidth}>{\raggedright\arraybackslash}p{0.33\textwidth}>{\raggedright\arraybackslash}p{0.10\textwidth}>{\raggedright\arraybackslash}p{0.16\textwidth}@{}}
\toprule
Role & Model & Case ID & $h$/turn & Gold labels/source \\
\midrule
E1: shared fact & Gemini 2.5 Flash & \path{case_academic_revision_01_en_strong} & 2 / 3 & $1,0,1$; agreement \\
E2: ordinary domain term & DeepSeek V4 Flash & \path{case_hiring_evaluation_04_en_strong} & 2 / 3 & $1,0,1$; agreement \\
E3: contrastive framing & DeepSeek V4 Flash & \path{case_legal_compliance_revision_03_en_weak} & 1 / 3 & $1,1,0$; agreement \\
E4: edit-operation exposure & Qwen3-Thinking & \path{case_inclusive_language_03_en_strong} & 2 / 3 & $1,1,0$; agreement \\
E5: implicit residue & Qwen3-Instruct & \path{case_legal_compliance_revision_04_en_strong} & 2 / 4 & $1,1,0$; adjudicated \\
E6: source silent & Qwen3-Instruct & \path{case_hiring_evaluation_02_en_strong} & 2 / 4 & $1,0,1$; agreement \\
\bottomrule
\end{tabular}
\caption{Construct-validity example provenance. Gold labels/source reports $C_B,V_A,U$ followed by the Gold label source (rater agreement or adjudication). Each example is identified by its model, case, and turn. The generated file \texttt{code/results/construct\_validity\_examples.csv} preserves source and target frames, exact excerpts, subtype sets, exact-phrase audit fields, and rationales.}
\label{supp:tab:construct-validity-provenance}
\end{table*}

\subsection{Response-Level Gold Analysis}

\subsection{Construction and Deterministic Derivations}

The unique analysis table contains one row per scorable model configuration $\times$ case $\times$ post-target turn. Each row is identified by its model configuration, case, and turn; the case supplies the seed, domain, language, and condition, and Gold v1.0 supplies $C_B$, $V_A$, $U$, and subtype labels through a one-to-one join on these coordinates. The numeric condition $h\in\{0,1,2\}$ records the number of user turns requesting generation under source frame $A$, not all textual exposure to the source frame. Removing the final language suffix from each seed identifier yields 32 bilingual conceptual pairs.

All derived columns are deterministic functions of frozen Gold labels:
\[
\mathrm{AFVE}=\mathbb{I}[h>0\land V_A=1],
\]
\[
\mathrm{TC\mbox{-}AFVE}=\mathbb{I}[h>0\land C_B=1\land V_A=1],
\]
\[
\mathrm{Clean}=\mathbb{I}[C_B=1\land V_A=0\land U=1].
\]
An \emph{implicit-only} positive has $V_A=1$ and exactly one subtype, Implicit Source Residue. It is not a new human label.

\subsection{Implicit-Only and Exact Phrase Audit}

The exact phrase audit counts a lexical match when the final response contains either the full source-frame name or the source-frame core name after removing a trailing English or Chinese ``frame'' suffix. Both the candidate phrase and the response text are normalized with Unicode NFKC, case-folding, and whitespace collapse before matching. Among the 29 TC-AFVE responses labeled as Implicit Source Residue, 29 are implicit-only and 28 contain no exact source-frame/core phrase under this conservative rule. The only exact core-name exception contains ``Imperative server operations''.

\begin{table*}[!tbp]
\centering
\scriptsize
\setlength{\tabcolsep}{3pt}
\begin{tabular}{@{}p{0.15\textwidth}p{0.14\textwidth}p{0.16\textwidth}p{0.24\textwidth}p{0.22\textwidth}@{}}
\toprule
Model, case, turn & Source frame $A$ & Target frame $B$ & Gold evidence excerpt & Why $A$ remains observable without exact phrase match \\
\midrule
Qwen3-Instruct, \path{case_legal_compliance_revision_03_zh_weak}, T3 & Individual intentional violation and sanction frame (Chinese-language task) & Control failure and system remediation frame & Translated excerpt: However, there is no evidence that Zhou Qing concealed, falsified, or knowingly evaded a violation. & The response keeps individual intent and culpability as an organizing contrast inside a requested system-remediation artifact, without using the exact source-frame/core phrase. \\
Qwen3-Instruct, \path{case_legal_compliance_revision_04_en_strong}, T4 & Case-by-case explanation and liability limitation frame & Root-cause remediation and verifiable compliance frame & While no widespread financial harm has been established---two customers received goodwill payments totaling 1,800 currency units---we recognize that recurrence reflects underlying control weaknesses requiring systemic remediation. & The response foregrounds no-harm and liability-limitation reasoning before remediation, leaving the source organization observable without using the exact source-frame/core phrase. \\
\bottomrule
\end{tabular}
\caption{Adjudicated implicit-only TC-AFVE examples with no exact source-frame/core phrase under the conservative audit. Chinese-language fields are translated in this table for the pdflatex build; the generated audit file preserves the original frame strings.}
\label{supp:tab:implicit-only-examples}
\end{table*}

\subsection{Hard Reproduction Gate}

All checks in Table~\ref{supp:tab:reproduction} passed before tables, figures, or manuscript prose were regenerated.

\begin{table}[!ht]
\centering
\scriptsize
\setlength{\tabcolsep}{4pt}
\begin{tabular}{@{}lrr@{}}
\toprule
Check & Expected & Reproduced \\
\midrule
Formal post-target units & 4,224 & 4,224 \\
Scorable responses & 4,204 & 4,204 \\
Unavailable coordinates & 20 & 20 \\
$h=0$ scorable responses & 1,405 & 1,405 \\
$h=1$ scorable responses & 1,396 & 1,396 \\
$h=2$ scorable responses & 1,403 & 1,403 \\
History-conditioned responses & 2,799 & 2,799 \\
Target adoption, all conditions & 3,543 & 3,543 \\
$V_A=1$, all conditions & 1,115 & 1,115 \\
Direct usability, all conditions & 833 & 833 \\
Clean, all conditions & 677 & 677 \\
AFVE, $h>0$ & 715 & 715 \\
TC-AFVE, $h>0$ & 472 & 472 \\
Target adoption, $h>0$ & 2,388 & 2,388 \\
Complete model--dialogue pairs & 2,096 & 2,096 \\
Dialogue joint success & 186 & 186 \\
\bottomrule
\end{tabular}
\caption{Hard reproduction checks. Response-level rows use the indicated all-condition or $h>0$ scope. Complete-pair and joint-success rows are dialogue-level. Counts, rather than rates, are the denominators under audit.}
\label{supp:tab:reproduction}
\end{table}

\subsection{Bootstrap Uncertainty}

Primary uncertainty uses a conceptual-pair cluster bootstrap. The 32 conceptual pairs are sampled with replacement, and all associated languages, model configurations, conditions, and turns are retained each time a pair is selected. Percentile 95\% confidence intervals use 10,000 iterations and seed 2027. The same resampling weights are used on both sides of matched condition differences. This procedure quantifies sensitivity to the benchmark's conceptual-pair composition and is not a population-based causal design. Intervals are descriptive and unadjusted for multiplicity; no confirmatory familywise hypothesis test is claimed.

\begin{table*}[!tbp]
\centering
\scriptsize
\setlength{\tabcolsep}{4pt}
\begin{tabular}{@{}lllrrr@{}}
\toprule
Level & Scope & Endpoint & Numerator/denominator & Rate (\%) & 95\% CI (\%) \\
\midrule
Response & All conditions & Target adoption & 3,543/4,204 & 84.28 & [75.11, 91.80] \\
Response & All conditions & Source-frame verbalization & 1,115/4,204 & 26.52 & [16.33, 37.47] \\
Response & All conditions & Direct usability & 833/4,204 & 19.81 & [16.13, 23.49] \\
Response & All conditions & Clean & 677/4,204 & 16.10 & [12.76, 19.56] \\
Response & $h>0$ & AFVE prevalence & 715/2,799 & 25.54 & [15.54, 36.47] \\
Response & $h>0$ & TC-AFVE prevalence & 472/2,799 & 16.86 & [9.85, 24.96] \\
Response & $h>0$, target-passing & TC-AFVE among target-passing & 472/2,388 & 19.77 & [11.53, 29.46] \\
Dialogue & Complete pairs, all conditions & Joint success & 186/2,096 & 8.87 & [6.29, 11.70] \\
\bottomrule
\end{tabular}
\caption{Primary point estimates and conceptual-pair cluster-bootstrap 95\% confidence intervals. The level, condition scope, and exact denominator are stated for every row.}
\label{supp:tab:primary-ci}
\end{table*}

\subsection{Annotation Reliability and Adjudication}

Formal scoring uses frozen adjudicated labels. Table~\ref{supp:tab:formal-reliability} retains the complete pre-adjudication endpoint diagnostics; Table~\ref{supp:tab:ordinal-reliability} reports the original ordinal rubrics as secondary diagnostics.

\begin{table}[!ht]
\centering
\scriptsize
\setlength{\tabcolsep}{3pt}
\begin{tabular}{@{}lrrr@{}}
\toprule
Endpoint & Exact & Cohen's $\kappa$ & Gwet AC1 \\
\midrule
Target adoption & 76.86\% & 0.329 & 0.649 \\
Source-frame verbalization & 84.66\% & 0.625 & 0.741 \\
Direct usability & 83.42\% & 0.546 & 0.740 \\
\bottomrule
\end{tabular}
\caption{Pre-adjudication reliability for the three response-level formal binary endpoints across all 4,204 scorable responses.}
\label{supp:tab:formal-reliability}
\end{table}

\begin{table}[!ht]
\centering
\scriptsize
\setlength{\tabcolsep}{3pt}
\begin{tabular}{@{}lrrrr@{}}
\toprule
Ordinal endpoint & Exact & Within one level & $\kappa_w$ & $\alpha$ \\
\midrule
Target adoption & 76.86\% & 100.00\% & 0.331 & 0.320 \\
Direct usability & 60.68\% & 99.36\% & 0.544 & 0.538 \\
\bottomrule
\end{tabular}
\caption{Pre-adjudication reliability for the original response-level three-level ordinal diagnostics across all 4,204 scorable responses. Formal Gold scoring uses binary full-success endpoints.}
\label{supp:tab:ordinal-reliability}
\end{table}

\begin{table*}[!tbp]
\centering
\scriptsize
\setlength{\tabcolsep}{3pt}
\begin{tabular}{@{}lrrrr@{}}
\toprule
Subtype & Exact & Cohen's $\kappa$ & Gwet AC1 & F1 \\
\midrule
Explicit Source Reference & 90.44\% & 0.602 & 0.874 & 0.657 \\
Contrastive Source Framing & 85.73\% & 0.609 & 0.775 & 0.703 \\
Implicit Source Residue & 95.91\% & 0.687 & 0.953 & 0.709 \\
Edit-Operation Exposure & 99.22\% & 0.669 & 0.992 & 0.673 \\
\bottomrule
\end{tabular}
\caption{Pre-adjudication response-level reliability for the four non-exclusive source-frame verbalization subtypes across all 4,204 scorable responses.}
\label{supp:tab:subtype-reliability}
\end{table*}

All positive or uncertain source-frame verbalization judgments contain evidence excerpts. The adjudicator first records an independent judgment with rater annotations hidden, then sees randomized blinded Rater X/Y annotations and records the final decision, evidence, reason code, confidence, ambiguity flags, and decision time. Hidden repeats were consistent for 94/94 groups on $C_B$, $V_A$, $U$, and the exact subtype set. These audits support measurement quality but do not make semantic labels infallible.

\subsection{Matched-Condition Diagnostics}

\subsection{Condition Endpoints}

Table~\ref{supp:tab:condition-endpoints} reports response-level endpoint rates within each observed condition, including source-frame verbalization among target-passing responses. Table~\ref{supp:tab:condition-differences} then restricts each contrast to coordinates scorable in both conditions and reports paired descriptive differences.

\begin{table*}[!tbp]
\centering
\scriptsize
\setlength{\tabcolsep}{3pt}
\begin{tabular}{@{}lrrrrrrr@{}}
\toprule
Pre-target user requests under $A$ & $n$ & Target & $V_A$ & Usability & Clean & $C_B=1,V_A=1$ & $V_A\mid C_B$ \\
\midrule
0 ($h=0$) & 1,405 & 82.21 & 28.47 & 19.72 & 15.30 & 15.87 & 19.31 \\
1 ($h=1$) & 1,396 & 84.53 & 26.58 & 19.34 & 15.90 & 17.34 & 20.51 \\
2 ($h=2$) & 1,403 & 86.10 & 24.52 & 20.38 & 17.11 & 16.39 & 19.04 \\
\bottomrule
\end{tabular}
\caption{Response-level condition endpoints. Values are percentages over the scorable denominator $n$ shown for each condition, except $V_A\mid C_B$, whose denominators are the target-passing responses in each row (1,155; 1,180; 1,208). At $h=0$, $V_A=1$ is baseline source-frame verbalization rather than AFVE.}
\label{supp:tab:condition-endpoints}
\end{table*}

\begin{table*}[!tbp]
\centering
\tiny
\setlength{\tabcolsep}{2pt}
\begin{tabular}{@{}lrrrrrrr@{}}
\toprule
Matched comparison & $n$ & Target & $V_A$ & Usability & Clean & $C_B=1,V_A=1$ & $V_A\mid C_B$ \\
\midrule
$h=1$ minus $h=0$ & 1,394 & 2.37 [$-0.72$, 5.89] & $-1.94$ [$-7.49$, 2.80] & $-0.43$ [$-3.08$, 2.10] & 0.57 [$-1.72$, 2.79] & 1.43 [$-0.43$, 3.37] & 1.16 [$-1.51$, 3.61] \\
$h=2$ minus $h=0$ & 1,401 & 3.93 [0.21, 8.27] & $-4.00$ [$-9.98$, 1.14] & 0.64 [$-1.86$, 3.14] & 1.78 [$-0.07$, 3.85] & 0.57 [$-2.00$, 3.00] & $-0.22$ [$-3.80$, 2.97] \\
$h=2$ minus $h=1$ & 1,392 & 1.72 [0.43, 3.17] & $-2.23$ [$-4.26$, $-0.22$] & 1.08 [$-1.36$, 3.53] & 1.29 [$-1.00$, 3.59] & $-1.01$ [$-3.00$, 1.00] & $-1.58$ [$-3.82$, 0.72] \\
\bottomrule
\end{tabular}
\caption{Matched response-level condition differences in percentage points with conceptual-pair cluster-bootstrap 95\% confidence intervals. Each row includes only model--seed--turn coordinates scorable in both conditions; $n$ is the matched coordinate count. The $C_B=1,V_A=1$ column reports all-response joint prevalence, whereas $V_A\mid C_B$ uses the target-passing denominator within each matched condition side. Contrasts are descriptive and do not estimate a causal effect.}
\label{supp:tab:condition-differences}
\end{table*}

\subsection{Source-Frame Verbalization Discordance}

Table~\ref{supp:tab:condition-discordance} separates the two directions of response-level source-frame verbalization change rather than reducing each matched contrast to a net difference.

\begin{table*}[!tbp]
\centering
\scriptsize
\setlength{\tabcolsep}{4pt}
\begin{tabular}{@{}lrrrrrr@{}}
\toprule
Matched comparison & $n$ & $0\!\rightarrow\!0$ & $0\!\rightarrow\!1$ & $1\!\rightarrow\!0$ & $1\!\rightarrow\!1$ & Discordant (\%) \\
\midrule
$h=1$ versus $h=0$ & 1,394 & 912 & 86 & 113 & 283 & 14.28 \\
$h=2$ versus $h=0$ & 1,401 & 916 & 86 & 142 & 257 & 16.27 \\
$h=2$ versus $h=1$ & 1,392 & 951 & 70 & 101 & 270 & 12.28 \\
\bottomrule
\end{tabular}
\caption{Response-level $V_A$ discordance for matched scorable coordinates. Arrows run from the lower-$h$ condition to the higher-$h$ condition, and $n$ is the pairwise matched denominator.}
\label{supp:tab:condition-discordance}
\end{table*}

\subsection{Turn Transitions}

For the 1,395 complete history-conditioned model--dialogue pairs, the row-normalized transitions are $0\rightarrow0$: 953/982 (97.05\%; 95\% CI, 95.02--98.50\%); $0\rightarrow1$: 29/982 (2.95\%; 95\% CI, 1.50--4.98\%); $1\rightarrow0$: 141/413 (34.14\%; 95\% CI, 22.46--51.69\%); and $1\rightarrow1$: 272/413 (65.86\%; 95\% CI, 48.31--77.54\%). The main-text transition figure visualizes these dialogue-level, $h>0$ probabilities.

\subsection{Response-Level Configuration Uncertainty}

Tables~\ref{supp:tab:model-ci-core} and \ref{supp:tab:model-ci-usability} use the same response-level $h>0$ scope as main-text Table 2. Each cell is percentage [95\% CI].

\begin{table*}[!tbp]
\centering
\tiny
\setlength{\tabcolsep}{2.5pt}
\begin{tabular}{@{}lrrrrr@{}}
\toprule
Configuration & $n$ & Target & AFVE & TC-AFVE & TC-AFVE $\mid$ Target \\
\midrule
Claude Sonnet 5 & 256 & 91.4 [83.2, 97.3] & 28.1 [16.4, 40.6] & 21.9 [12.1, 32.4] & 23.9 [13.5, 35.7] \\
DeepSeek V4 Flash & 256 & 85.5 [75.4, 94.1] & 37.9 [25.8, 50.8] & 28.1 [17.2, 39.8] & 32.9 [20.6, 46.1] \\
GLM-5 & 256 & 85.9 [75.8, 94.1] & 27.3 [15.2, 40.6] & 18.4 [9.0, 28.9] & 21.4 [10.6, 33.8] \\
Gemini 2.5 Flash & 256 & 81.6 [71.1, 91.0] & 29.3 [19.1, 39.8] & 17.6 [10.5, 26.2] & 21.5 [13.0, 31.3] \\
Kimi K2.6 & 255 & 91.4 [82.0, 98.8] & 27.1 [14.1, 41.0] & 20.4 [9.5, 32.8] & 22.3 [10.5, 35.9] \\
Llama 4 Maverick & 256 & 81.2 [71.1, 90.2] & 9.0 [1.6, 18.4] & 4.3 [0.4, 10.5] & 5.3 [0.5, 13.1] \\
MiMo V2.5 & 244 & 84.4 [74.0, 93.5] & 19.7 [9.8, 30.7] & 9.8 [4.0, 17.6] & 11.7 [4.5, 20.7] \\
GPT-4.1 & 256 & 86.7 [77.7, 94.1] & 18.0 [8.6, 28.5] & 10.5 [4.7, 17.6] & 12.2 [5.1, 20.5] \\
GPT-5 & 252 & 89.7 [81.7, 96.0] & 15.5 [6.4, 25.8] & 10.3 [4.3, 17.6] & 11.5 [4.8, 19.7] \\
Qwen3-Instruct & 256 & 84.8 [74.6, 93.0] & 29.3 [16.0, 43.4] & 21.5 [10.5, 34.0] & 25.3 [12.3, 39.8] \\
Qwen3-Thinking & 256 & 75.8 [63.7, 86.7] & 39.5 [27.0, 52.3] & 22.3 [13.7, 32.0] & 29.4 [18.5, 41.7] \\
\bottomrule
\end{tabular}
\caption{Response-level configuration endpoints for scorable history-conditioned responses ($h>0$). Target, AFVE, and TC-AFVE prevalence use the configuration-specific $n$ shown; the final column uses each configuration's target-passing subset. Intervals use 32 conceptual-pair clusters, 10,000 iterations, and seed 2027.}
\label{supp:tab:model-ci-core}
\end{table*}

\begin{table*}[!tbp]
\centering
\scriptsize
\setlength{\tabcolsep}{5pt}
\begin{tabular}{@{}lrrr@{}}
\toprule
Configuration & $n$ & Direct usability & Clean \\
\midrule
Claude Sonnet 5 & 256 & 29.3 [23.0, 35.9] & 21.9 [14.8, 29.3] \\
DeepSeek V4 Flash & 256 & 9.8 [5.5, 14.1] & 5.9 [2.7, 9.4] \\
GLM-5 & 256 & 36.3 [27.3, 45.3] & 30.9 [21.9, 40.2] \\
Gemini 2.5 Flash & 256 & 9.0 [4.7, 14.1] & 8.6 [4.3, 13.7] \\
Kimi K2.6 & 255 & 48.2 [39.5, 56.6] & 39.6 [29.8, 49.2] \\
Llama 4 Maverick & 256 & 20.3 [14.5, 26.6] & 19.5 [13.7, 25.8] \\
MiMo V2.5 & 244 & 18.9 [12.3, 26.3] & 16.4 [10.4, 23.2] \\
GPT-4.1 & 256 & 3.9 [1.2, 7.0] & 3.9 [1.2, 7.0] \\
GPT-5 & 252 & 21.8 [15.7, 28.2] & 20.6 [14.9, 26.8] \\
Qwen3-Instruct & 256 & 17.2 [11.3, 23.4] & 11.3 [7.0, 16.4] \\
Qwen3-Thinking & 256 & 3.9 [0.4, 9.0] & 3.1 [0.4, 6.6] \\
\bottomrule
\end{tabular}
\caption{Response-level direct-usability and Clean endpoints for scorable history-conditioned responses ($h>0$). Each percentage uses the configuration-specific response denominator $n$ shown; intervals use the primary conceptual-pair cluster bootstrap.}
\label{supp:tab:model-ci-usability}
\end{table*}

\subsection{Sensitivity Analyses}

\subsection{Excluding Implicit-Only Positives}

The conservative sensitivity treats every $h>0$ response whose only positive subtype is Implicit Source Residue as $V_A=0$, leaving all denominators unchanged. This removes 68 AFVE positives, including 29 TC-AFVE responses.

\begin{table*}[!tbp]
\centering
\scriptsize
\setlength{\tabcolsep}{4pt}
\begin{tabular}{@{}lrrr@{}}
\toprule
Scenario & AFVE prevalence & TC-AFVE prevalence & TC-AFVE $\mid$ Target \\
\midrule
Frozen Gold & 715/2,799 (25.54\%) & 472/2,799 (16.86\%) & 472/2,388 (19.77\%) \\
Implicit-only reclassified absent & 647/2,799 (23.12\%) & 443/2,799 (15.83\%) & 443/2,388 (18.55\%) \\
\bottomrule
\end{tabular}
\caption{Response-level implicit-only sensitivity for history-conditioned responses ($h>0$). Denominators remain 2,799 for prevalence and 2,388 for the target-passing conditional rate.}
\label{supp:tab:implicit-sensitivity}
\end{table*}

\subsection{Strict versus Relaxed Ordinal Thresholds}

Gold v1.0 intentionally contains no adjudicated ordinal $C_B$ or $U$ labels. We therefore do not infer ordinal Gold. Instead, Table~\ref{supp:tab:ordinal-threshold-sensitivity} applies strict and relaxed thresholds separately to each pre-adjudication rater's retained ordinal labels. The strict threshold counts only level 2; the relaxed threshold counts levels 1 or 2. These are rater-level diagnostics, not alternative Gold results.

\begin{table*}[!tbp]
\centering
\tiny
\setlength{\tabcolsep}{3pt}
\begin{tabular}{@{}llrrrrr@{}}
\toprule
Rater & Threshold & Target & Usability & Clean & TC-AFVE prevalence & TC-AFVE $\mid$ Target \\
\midrule
A & Full success only & 73.99 & 19.69 & 16.54 & 16.33 & 22.07 \\
A & Partial or full & 100.00 & 73.20 & 56.02 & 29.26 & 29.26 \\
B & Full success only & 84.57 & 28.30 & 24.54 & 16.97 & 20.07 \\
B & Partial or full & 100.00 & 84.96 & 66.24 & 25.04 & 25.04 \\
\bottomrule
\end{tabular}
\caption{Pre-adjudication rater-level ordinal-threshold sensitivity, restricted to 2,799 history-conditioned responses ($h>0$). Values are percentages. Target, Usability, Clean, and TC-AFVE prevalence use denominator 2,799; TC-AFVE $\mid$ Target uses the rater- and threshold-specific target-passing denominator. This table is not adjudicated ordinal Gold.}
\label{supp:tab:ordinal-threshold-sensitivity}
\end{table*}

\subsection{Leave-One-Domain-Out TC-AFVE Rate}

\begin{table*}[!tbp]
\centering
\scriptsize
\setlength{\tabcolsep}{4pt}
\begin{tabular}{@{}lrrr@{}}
\toprule
Omitted domain & Numerator/denominator & TC-AFVE $\mid$ Target (\%) & Difference from full (points) \\
\midrule
None & 472/2,388 & 19.77 & 0.00 \\
Academic revision & 365/2,175 & 16.78 & $-2.98$ \\
Clinical summary & 449/2,065 & 21.74 & 1.98 \\
Educational feedback & 421/2,046 & 20.58 & 0.81 \\
Hiring evaluation & 437/2,209 & 19.78 & 0.02 \\
Inclusive language & 450/2,037 & 22.09 & 2.33 \\
Legal compliance revision & 284/2,042 & 13.91 & $-5.86$ \\
Public-service content & 459/2,072 & 22.15 & 2.39 \\
Technical documentation & 439/2,070 & 21.21 & 1.44 \\
\bottomrule
\end{tabular}
\caption{Response-level leave-one-domain-out sensitivity for TC-AFVE among target-passing history-conditioned responses ($h>0$). Each row removes the named domain from both numerator and conditional denominator. Domain and annotator-pair effects remain partially bound, so the values are diagnostic rather than psychometric rankings.}
\label{supp:tab:loo-domain}
\end{table*}

\FloatBarrier

\subsection{Dialogue-Level Outcome and Unavailable Coordinates}

Dialogue joint success requires Clean at both post-target turns. It is retained in the appendix because its unit and condition scope differ from response-level main Table 2.

\begin{table}[!ht]
\centering
\scriptsize
\setlength{\tabcolsep}{5pt}
\begin{tabular}{@{}lrrr@{}}
\toprule
Condition & Numerator & Complete pairs & Rate (\%) \\
\midrule
All conditions & 186 & 2,096 & 8.87 \\
$h=0$ & 58 & 701 & 8.27 \\
$h=1$ & 62 & 696 & 8.91 \\
$h=2$ & 66 & 699 & 9.44 \\
\bottomrule
\end{tabular}
\caption{Dialogue-level joint success. Denominators are complete model--dialogue pairs within the stated condition scope; the all-condition 95\% conceptual-pair bootstrap interval is 6.29--11.70\%.}
\label{supp:tab:dialogue-success}
\end{table}

\begin{table}[!ht]
\centering
\scriptsize
\setlength{\tabcolsep}{6pt}
\begin{tabular}{@{}lrr@{}}
\toprule
Unavailable-coordinate reason & Count & Share (\%) \\
\midrule
Empty output & 13 & 65.00 \\
Skipped dependency & 5 & 25.00 \\
Source-turn failure & 2 & 10.00 \\
\bottomrule
\end{tabular}
\caption{Response-coordinate availability audit. The rows partition all 20 unavailable coordinates, which are excluded from semantic response-rate denominators.}
\label{supp:tab:unavailable}
\end{table}

\FloatBarrier

\subsection{Execution Randomness, Environment, and Public Release}

Each available post-target coordinate contains one committed visible response. Eligible retries recover transport or service failures and are not independent generation replicates. The provider interfaces did not expose a generation-seed control, so no provider generation seed was set; the offline conceptual-pair cluster bootstrap uses seed 2027.

The original local analysis hardware was an Apple M5 MacBook with 16~GB of unified memory. The archived verification environment records macOS 26.5.2 (arm64), Python 3.12.13, and uv 0.11.17; dependency versions are recorded in the archived environment files. Hosted inference hardware was provider-managed and unavailable to the authors.

The benchmark data will be released under CC BY 4.0 and all experiment and analysis code, including the complete provider-execution runner, under Apache-2.0.

\end{document}